# Techniques Toward Optimizing Viewability in RTB Ad Campaigns Using Reinforcement Learning

Michael Tashman, John Hoffman, Jiayi Xie, Fengdan Ye[1], Atefeh Morsali[1], Lee Winikor, Rouzbeh Gerami

Copilot AI at Xaxis
New York, USA
{michael.tashman, john.hoffman, jiayi.xie, lee.winikor, rouzbeh.gerami}@xaxis.com

## ABSTRACT

Reinforcement learning (RL) is an effective technique for training decision-making agents through interactions with their environment. The advent of deep learning has been associated with highly notable successes with sequential decision making problems — such as defeating some of the highest-ranked human players at Go. In digital advertising, real-time bidding (RTB) is a common method of allocating advertising inventory through real-time auctions. Bidding strategies need to incorporate logic for dynamically adjusting parameters in order to deliver pre-assigned campaign goals. Here we discuss techniques toward using RL to train bidding agents. As a campaign metric we particularly focused on viewability: the percentage of inventory which goes on to be viewed by an end user.

This paper is presented as a survey of techniques and experiments which we developed through the course of this research. We discuss expanding our training data to include edge cases by training on simulated interactions. We discuss the experimental results comparing the performance of several promising RL algorithms, and an approach to hyperparameter optimization of an actor/critic training pipeline through Bayesian optimization. Finally, we present live-traffic tests of some of our RL agents against a rule-based feedback-control approach, demonstrating the potential for this method as well as areas for further improvement. This paper therefore presents an arrangement of our findings in this quickly developing field, and ways that it can be applied to an RTB use case.

## 1. INTRODUCTION

In order to purchase ad inventory in Real-Time Bidding markets, bidding strategies should be employed that aim to balance different measures of success of an advertising campaign. At the core of such bidding strategies are machine learning models that predict the probability of user response to an ad impression. The bid price is generally a non-decreasing function of this probability, whose exact shape reflects the balance of different KPIs—Key Performance Indicators—designed to measure the effectiveness of the bidding strategy. Some KPIs are constructed against a specific target goal: for example, the *cost per action* of an ad campaign is directly linked to the advertiser's return on investment. Other KPIs reflect the health of a campaign—such as *pacing*, which ensures that a campaign is on track to smoothly spend its full budget by the time the campaign ends. Finally, some KPIs are used as heuristics for the kind of inventory being purchased. *Viewability* measures the percentage of purchased inventory that goes on to be visible on the user's display. A poor viewability rate implies that the bidding strategy is not choosing the best inventory, even if all of the other KPIs appear reasonable, because the ad cannot have impact if it was not viewed.

The optimal bid function should allow optimization of different combination of the KPI goals and can be parameterized to reflect this. An effective bidding strategy needs a dynamic updating system to continuously adjust such bid parameters. For this, classic feedback control methods have been proven successful. Tashman, et al [2020] describes such a system, including a controller with a PID module, an actuator to take action according to the output of the controller, and heuristics rules on how to take actions. For example, the actuator was provided with several "levers" (such as an ability to apply bid multipliers and no-bid criteria) and programmed with the relationship between each lever and its expected directional effect on each KPI. Through small, continuous adjustments, the feedback control loop proved highly effective in moving the overall system toward increasingly optimal states.

Reinforcement learning has proven to be an effective method of machine decision-making. Unlike a rule-based feedback loop, RL is able to develop a nuanced interaction model based on real-world encounters. It is capable of identifying sophisticated and non-obvious methods of achieving goals—as exemplified by its widely-heralded success in fields such as autonomous driving and playing Go [Silver et al., 2016]. With RL it is no longer necessary to take action based on pre-defined, heuristic rules, and the optimal policy can be learned by observing results of direct interactions with the environment.

---
[1] Work done during internship at Copilot AI.

In order to learn the optimal policy, RL requires sufficient data to learn generalizable rules. In many real-world contexts, including ours, it is not uncommon for this to be an insurmountable challenge. Therefore, our goal was to train an RL policy that would be capable of meaningfully improving campaign KPIs, while being trained from datasets that we would readily have available.

The long-term goal for this research is a trained agent that can handle a state space with many of the features common in RTB bidding, such as time of day, device type, and position on the page — and then output decisions for multiple possible actions. This would ideally include the ability to balance two or more KPI goals. Most of the research presented here is comprised of the starting phases of this project, in which we develop an agent for one KPI and one goal, with a relatively small state space.

In what follows we present research on training an RL agent for controlling viewability in an RTB ad campaign. The interaction model is relatively simple. In order to influence viewability, we use a *viewability threshold*: that is, given predictions of the view-probability of any unit of inventory under auction, we can choose not to place a bid in any auction in which this probability is too low. By moving a view-probability cutoff value (the viewability threshold) up or down, we can directly affect the overall viewability rate of the campaign. This interaction should be relatively straightforward to train, while promising significant benefit over a trial-and-error approach.

This paper discusses an assortment of our findings in the process of using reinforcement learning to control viewability: some effective methods of automating policy training through training pipelines; model-free vs. model-based training; relative performance of some popular RL algorithms; and performance compared with a rule-based feedback loop. This paper can be viewed as a survey of various RL algorithms and their tradeoffs, when applied to viewability control in RTB environment.

## 2. BACKGROUND

In this section, we will discuss relevant prior work. This largely covers three areas: research on using simpler feedback loops to control RTB campaigns, research on using RL for decision-making, and model-based and model-free RL.

### 2.1 Feedback Control Methods

When controlling an RTB campaign, we have several levers, such as bid multipliers and viewability threshold, with which to adjust the behavior of the campaign. Conversely, we have several KPIs, such as cost per action (CPA), cost per click (CPC), and viewability rate, with which to measure success. Because RTB bidding typically involves placing many bids throughout the day — and receiving results with a delay of no more than several hours — we have a short feedback loop, in which we can observe the results of behavioral changes in near-real time.

Several methods have been implemented to automate the optimization of RTB campaigns with a feedback-control approach. In Zhang et al. [2016], a *controller* module is paired with an *actuator*, in which the former assesses the state of the system, after which the latter takes action accordingly. This method was effective in improving cost per click, and validates that this is a reasonable approach.

In Gayek et al. [2016], authors introduce an algorithm to incrementally balance viewability with pacing. Their method requires predictions of the probability that an ad will be viewed, $p(v)$. Accordingly, one can set a viewability threshold, $\varphi(v)$, such that no bids will be submitted for any inventory where $p(v) < \varphi(v)$. It follows that adjusting $\varphi(v)$ will result in an adjustment in the final average viewability of the ad campaign. The algorithm proposed by Gayek et al., [2016] checks, at regular intervals, whether ad delivery is sufficient. If it is, $\varphi(v)$ is raised by a small increment; if not, $\varphi(v)$ is reduced. This method should eventually approximate $\varphi(v)^*$ — the highest viewability threshold which will not threaten delivery.

Kitts et al. [2017] proposes a multi-goal optimization approach built under the assumption that KPIs are not necessarily immutable constraints, but rather a general window into the performance of the campaign. By their reasoning, KPIs are simply a means by which stakeholder can assess the campaign's viability, and therefore a bidding agent should be robust to circumstances where not all KPIs can be satisfied at the same time. In this case, an agent should simply seek to achieve acceptable performance, while minimizing across-the-board KPI error as much as possible.

Tashman et al. [2020] proposes an RTB optimization model similar to Gayek et al. [2016], but extended to a larger number of KPIs. Their method associates each KPI with one or more levers to control it — such as bid price multiplier or viewability threshold — and uses a feedback loop with a PID module to throttle the lever. The KPIs to control are given an ordering by priority, in which a single KPI is selected to be optimized at a time.

Tashman et al. [2020] propose two methods for selecting the KPI to optimize. In "Simple Sequential" control, the higher-priority KPIs are favored in a similar manner as with Gayek et al. [2016], in which a KPI can only be selected for optimization if all KPIs with a higher priority are already in an acceptable range. In "Smart Sequential" control, the priority of each KPI is considered along with its distance from the relevant goal value, to create an overall score to quantify the degree to which the KPI is in need of optimization. The KPI in greatest need of optimization is selected in this way on a regular time interval.

## 2.2 Decision-Making with RL

Reinforcement learning has been established as a method for sequential decision-making for decades: its main distinction is its ability to learn through trial-and-error, interacting directly with its environment rather than building on prescribed rules. [Arulkumaran, et al., 2017]

Prior to the advent of deep learning, RL had been mostly effective in learning simple games and interaction models. The cartpole game — in which a simple physics simulator presents a pole balanced atop a cart, and the cart needs to be moved along a single axis to keep the poll from falling — was an early RL success. RL also proved effective at relatively simple card games such as blackjack. [Gullapalli, 1991; Perez-Uribe, 1998]

The development of deep learning dramatically increased the scope of RL's capabilities — enabling it to train effectively in situations with high-dimensional state or action spaces which would previously have been intractable [Arulkumaran et al., 2017]. Some of these use cases were revolutionary: an early deep RL success involved training a policy to play Atari 2600 games from only the pixels of the game output and the current score; the resulting agent played at a superhuman level [Minh et al., 2015].

Soon after, AlphaGo became noteworthy when it defeated the highest-ranked human player in Go. Unlike Chess, the action space in Go is too large to be feasibly mapped into a decision tree. Handcrafted rules, common in Chess simulators, are also not employed: AlphaGo directly learns the game through supervised and reinforcement learning [Silver et al., 2016; Arulkumaran et al., 2017].

Deep reinforcement learning has also been successfully applied in the RTB ad buying context. In Wu et al. [2018], researchers at Alibaba propose a reinforcement learning approach to buy RTB inventory in a second-price environment, with the need to adhere to a budget constraint. Authors discretize their action space, making it possible to use a Deep Q Network (DQN) algorithm; the balance of exploration vs. exploitation is handled through an adaptive ε-greedy approach. The action taken by the RL is the setting of a bid scaling factor, $\lambda$, that increased or decreased bids; the bids were calculated as $b = v/\lambda$, where $v$ is the expected value of the impression (e.g., Goal CPC * estimated CTR). This approach, having performed successfully, was used into production in Alibaba.

In Zhao et al. [2018], researchers (also at Alibaba) built a deep RL approach to buy RTB inventory in the sponsored search domain. Sponsored search tends to present a more difficult problem than display advertising, as multiple ads will typically win an auction and be presented to the user, with the winning ad simply getting the highest-rank placement. Optimizing sponsored search bidding therefore requires optimizing multiple bid prices for each ad — one per keyword — and it is thus insufficient to attempt to predict the simple market-clearing price for an auction.

The authors therefore propose a DQN to output a multiplier against the normally-calculated bid price for any given keyword. Recognizing the periodicity of the bidding environment on an hour-by-hour basis, authors train the RL policy against a dataset in which the hour of the day is one of the state features, and all auction statistics are aggregated to an hourly level. This method avoids the complexities of attempting to train a policy directly against impression-level data. Instead, a more traditional linear approximator approach is used to calculate impression-level bid prices, and the RL agent's output is simply used to scale the bids to account for daily market periodicity. (Authors refer to this approach as "control-by-model", in contrast with "control-by-action", in which an RL policy would directly choose bid prices.) This approach achieved strong performance and was deployed in a production environment at Alibaba.

## 2.3 Model-Based vs. Model-Free

Reinforcement learning training approaches can be split into two camps: those that employ a model to estimate transition probabilities directly ("model-based" methods) and those that learn policies that are agnostic to the transition dynamics ("model-free" methods). Model-based learning has the potential to make better use of limited available data by simulating the environment itself and imparting this knowledge to a model-free policy [Nagabandi et al., 2017; Ha and Schmidhuber 2018; Deisenroth and Rasmussen 2011].

[Nagabandi et al., 2017] showed that model-based learning does indeed provide a more data efficient means of achieving a baseline level of performance. However, because model-free methods are agnostic to the dynamics of their environment, they are able to achieve better asymptotic performance than model-based methods, which saturate performance once the modeling assumptions are no longer sufficiently accurate to describe the transition dynamics.

## 3. REINFORCEMENT LEARNING DESIGN

Here we describe the formulation of the RL problem. The state space contains three variables:
- Measured viewability at time $t$ ($v_t$)
- Viewability goal at time $t$ ($vg_t$)
- Viewability threshold set at time $t-1$ ($\varphi_{t-1}$)

Given that viewability measurements occur at discrete times, separated by a minimum of several hours, it is expected that a causal relationship will always exist between $\varphi_{t-1}$ and $v_t$.

The reward is a function of measured viewability and goal:

$$r_t = \left(1 - abs(v_t - vg_t)\right)^2 \quad (1)$$

The reward will always be in the range [0, 1], and higher values always indicate better performance. The value is squared, to slightly reduce the amount of reward gained as a function of increased viewability, thus raising the difficulty of the task. This exponent may be adjusted to increase or decrease difficulty as needed.

The action given by the policy is the new viewability threshold at time $t$: $\varphi_t$. This viewability threshold directly influences $v_{t+1}$.

When the agent is deployed against live traffic, the new viewability threshold is uploaded to our DSP, at which point it immediately adjusts our bidding logic. At the next measurement interval ($t+1$) we determine the new reward, $r_{t+1}$. This relationship is shown in Figure 2.

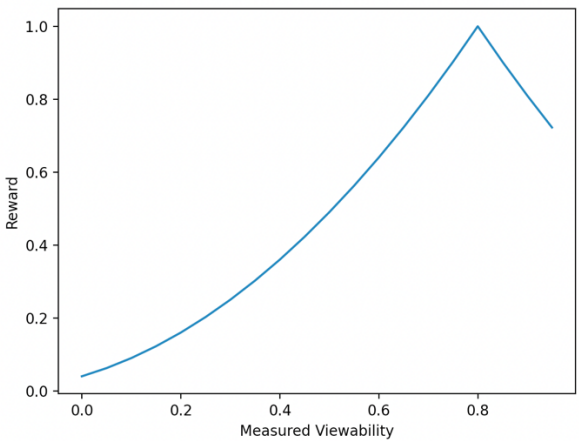

*Figure 1: Reward curve as a function of measured viewability, for a viewability goal of 80%.*

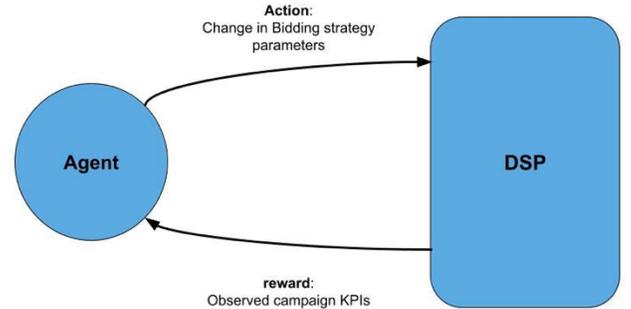

*Figure 2: The action-reward relationship between the RL agent and the DSP.*

## 4. DATA SOURCES

Training an effective reinforcement learning policy requires large amounts of data, even for relatively simple interactions. Because we buy RTB ad inventory in large quantities, we have a high volume of viable training data. We also developed several methods to extend this data, and to simulate unlikely scenarios.

### 4.1 Real-Traffic Data

Our most straightforward approach to gathering data is to use the aggregated records of the inventory that we have purchased. At regular intervals, we can monitor the average viewability of a campaign, along with the viewability threshold and viewability goal, and thereby train a policy.

### 4.2 Simulated Data with Stochastic Actions

Although the use of real-traffic data is a straightforward way to train a policy, it may lack sufficient volume and contain biases, particularly for edge cases. For instance, our bidding and campaign control systems will attempt to avoid undesirable scenarios as much as possible — such as setting a viewability threshold so high that delivery gets reduced to near-zero. However, in order to train an effective policy we need to be aware of these scenarios, in particular because we do not want the policy to simply mirror our current bidding strategies, but to improve upon them. To continue the example, we would prefer the RL agent to know when it would or would not be advantageous to raise the viewability threshold to normally dangerous levels; we can't realistically do this if we are training from the output of a bidding system that is specifically designed to avoid such situations. In order to rectify this, we built an RTB bidding environment simulator which outputs the predicted average viewability of a campaign for any input viewability threshold, including extreme values.

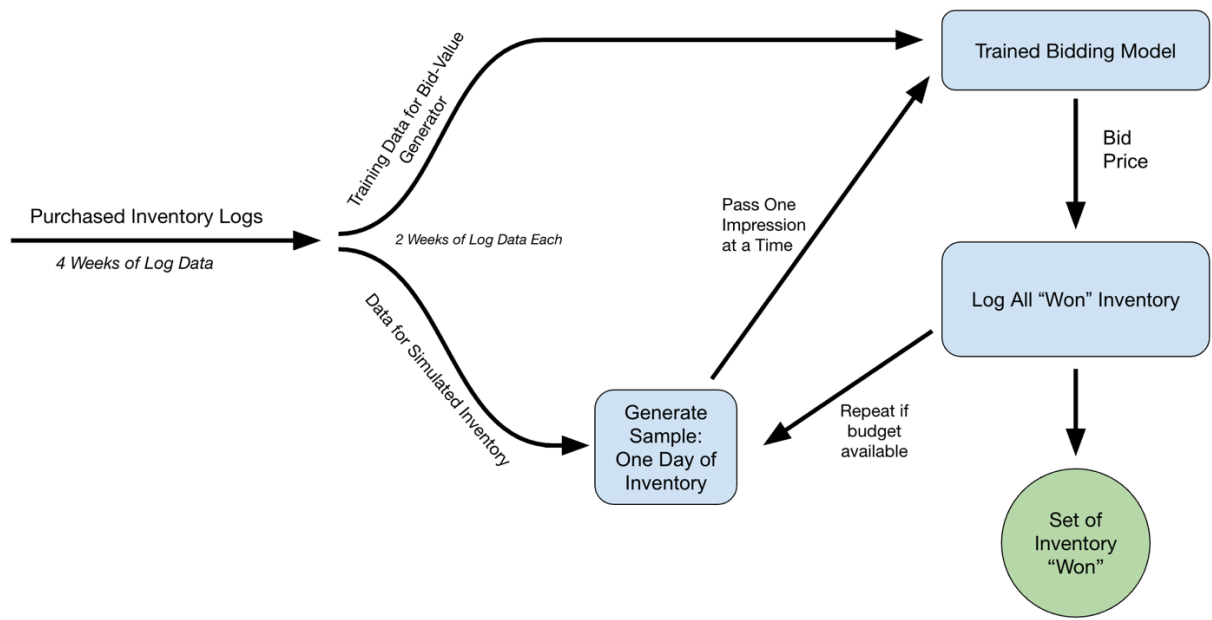

*Figure 3: Offline bidding simulator based on real-world purchased inventory logs.*

The RTB simulator takes as input a set of log-level data (LLD) — our most granular bidding logs, with one row per bought impression, including data such as domain, time-of-day, and device type. This also includes metrics such as the final cost of the impression, whether it was viewable, and whether it received a success event (typically a click or conversion).

With this data, we can run a simulated period of RTB ad-buying. We typically start with a month's worth of LLD data. With the first two weeks of data we train our production bid pricing model. (This uses a regression approach, trained on inventory features and success-event probability, to choose an optimal bid price for each auction.) We then randomly sample $n$ rows from the second two weeks of LLD, where $n$ is the approximate number of auctions in which we would participate during a typical day of bidding. We pass each impression, one at a time, into our production bidding algorithm, as if it was a live auction, to which it returns a bid price. Because most auctions are second-price, if the bidding algorithm returns a price greater than or equal to the actual price we paid for that impression, then we consider that to be a won auction; if the bid price is below our cost, we consider it lost. After all "bids" are placed, we can analyze the set of "won" bids, and assess whether we reached the viewability goal, cost goals, etc. In this way, we can simulate any viewability threshold values we want, including extreme values and uncommon transitions.

Note that there are some caveats to the method. First, in a real auction the DSP would provide us the probability, in their estimation, of that impression being viewable. In our simulator, although we do know the final result of whether the impression was viewable, we do not know the prediction that was originally passed through at bid time. To approximate this prediction, we built a logistic regression model, using all of the LLD features available to us, to predict each impression's view probability. For each simulated auction, we use this model to calculate the view probability, and pass it — along with all of the other LLD features — to our production bidding model.

The second issue is that our LLD only contains records for auctions that we have won; DSP reporting almost always excludes records for inventory that had been bid on and lost. This introduces a risk of bias, as it is not realistic to assume that our purchased impressions are fully generalizable to the inventory that we will have the opportunity to buy later. While we have not formally quantified the bias inherent to this method, we have carried out tests with both this method and against real-world traffic, such as in Tashman et al. [2020]. In these, the offline simulator provided results that were similar to those of the real-world environment, particularly in the interaction between the action taken by the feedback control system and the subsequent change in state space.

This suggests that the simulator output is able to serve as an effective approximation of real-world interactions.

## 5. ENVIRONMENT MODEL

An additional method of training a reinforcement learning model is to learn optimal policies from a trained environment model: a learned model which can return a subsequent state from a given state-action combination. This model-based approach can be an effective, efficient means of training, especially with smaller amounts of training data. (Conversely, model-free methods tend to be more effective when large sets of training data are available.)

Our model-based research is the focus of ongoing investigation; however we provide a simple model-based approach here to generate a baseline policy.

We consider the following model:

$$logit(v_{t+1}) = logit(v_t) + \alpha(logit(\varphi_{t+1}) - logit(\varphi_t)) \quad (2)$$

Where we have a standard logit function:

$$logit(x) = \log x/(1-x) \quad (3)$$

And $\varphi_t$ and $v_t$ are the viewability threshold and measured viewability at time $t$, respectively. Thus, our environment model $f$ to predict $v_{t+1}$ is

$$f(v_t, \varphi_t, \varphi_{t+1}) = \sigma(logit(v_t) + \alpha(logit(\varphi_{t+1}) - logit(\varphi_t))) \quad (4)$$

Note that $\sigma(x) = 1/(1 + e^{\wedge}(-x))$ is the sigmoid function. The transition probability implied by this model is therefore:

$$p(v_{t+1}|v_t, \varphi_t, \varphi_{t+1}) = \delta(v_{t+1} - f(v_t, \varphi_t, \varphi_{t+1})) \quad (5)$$

Where $\delta(x)$ is the Dirac delta function.

We expect that $v$ will be a monotonically increasing function of the viewability threshold; with all else being equal, a higher viewability threshold should yield higher viewability, and vice-versa. This means we need to choose a positive value of $\alpha$.

Our method for choosing this $\alpha$ was based on our past campaign data. For all datapoints in which we could pair viewability and view threshold for a given time $t$, with subsequent viewability and view threshold in time $t+1$, and for which the view threshold changed in between $t$ and $t+1$, we calculate:

$$\hat{\alpha} = \frac{logit(v_{t+1}) - logit(v_t)}{logit(\varphi_{t+1}) - logit(\varphi_t)} \quad (6)$$

We experimented with two different settings of $\alpha$:

$$\alpha = median(\hat{\alpha}) = 0.204$$

and

$$\alpha = mean(\{\hat{\alpha} \mid \hat{\alpha} > 0\}) = 1.08$$

### 5.1 Baseline Policy

Given a goal viewability $vg$, we can construct an objective reward function (Equation 1):

$$r(v) = (1 - abs(v - vg))^2$$

Note that it is important to penalize cases in which viewability is above- as well as below-goal. This is because viewability in excess of the goal is almost always achieved at some expense, whether in inventory cost or in our ability to carry out smooth spending.

For our reward function, we presume that the optimal policy under the baseline model is greedy:

$$\pi^*(v_t, \varphi_t, \varphi_{t+1}) = argmax_{\varphi_{t+1}} r(f(v_t, \varphi_t, \varphi_{t+1})) \quad (7)$$

Since, for any desired $vg \in (0, 1)$ and from any desired $(v_t \in (0, 1), \varphi_t \in (0, 1))$, we are able to select a subsequent viewability threshold $\varphi_{t+1}$ that will provide us the maximum possible reward at that point. This allows us to maximize next-step rewards, and discount rewards further in the future.

## 6. METHODOLOGY

### 6.1 Horizon / ReAgent

A large portion of our training was done with Horizon/ReAgent, a deep reinforcement learning package developed by Facebook specifically for training against offline datasets. [Gauci et al., 2018] (This package was initially named Horizon, but was later renamed to ReAgent, which is how we will refer to it.)

ReAgent is built on PyTorch, and is designed for quickly training and deploying RL policies in production systems. Its notable improvement over other packages is that it was specifically designed for off-policy training on previously logged interactions, without the need to interactively explore the action space in a simulator or live environment. It includes a Spark pipeline to help parallelize the process of converting large quantities of data into its internal format. ReAgent supports several popular algorithms, such as Deep Q-Network (DQN), Deep Deterministic Policy Gradient (DDPG) and Soft Actor-Critic (SAC). Finally, it includes a number of diagnostic tools in a built-in dashboard.

We used ReAgent for most of our training, particularly DQN and DDPG. It was helpful to make some additions to its processes, however. The Spark training pipeline had too much overhead for our needs, so we built a lightweight Python pipeline to carry out the data conversion process and return identical output.

Furthermore, we needed a fast way to verify whether a policy was effective, beyond the diagnostics and before running a time-consuming bid simulation (Section 4.2). To do this, we built a set of tests that would quickly run on a trained policy, and could be set to run automatically. This includes a simple toy environment, in which an agent is tasked with adjusting viewability in a roughly correct manner to meet a simple goal; the module records whether the agent is able to meet the goal, and if so, the number of intervals required. (The toy model's logic, converting a viewability threshold into a new "observed" viewability, is a very naive linear function which can run in milliseconds. It is intended to only very roughly simulate a real environment.)

A second test generates a toy environment in which the viewability is already at goal, and verifies that the agent does not attempt to change the threshold meaningfully. The final test challenges the agent with an assortment of common environment variable values. It then lists the results, so that an observer can tell, at a glance, whether the agent is capable of performing rationally. Although this method does not return a single objective score to quantify performance, it proved useful for catching ineffective policies early.

## 6.2 Algorithm Choice

One of the fundamental decisions in training a reinforcement learning policy is that of choosing which particular training algorithm to use. A variety of choices exist, each associated with a set of advantages and disadvantages.

The **Discrete Action Deep Q-Network** (DQN) is a common and relatively simple approach which is typically associated with state-of-the-art performance. [Gauci et al. 2018; Minh et al., 2015] The primary disadvantage of this approach is the discrete action space. For our purposes, the viewability threshold is best approached as a continuous space; in order to use a DQN, we needed to group all possible actions into discrete bins. Thus, part of our research was in comparing the effectiveness of different numbers of bins when training a DQN.

The **Deep Deterministic Policy Gradient** (DDPG), by contrast, is designed to work with a continuous action space, avoiding any issue with binning the action space [Lillicrap et al., 2015]. However, this algorithm requires two functions — both policy ("actor") and value ("critic") models must be trained. This requires two sets of training hyperparameters. In our experience with Bayesian optimization (Section 7), effective policies frequently had different hyperparameters between actor and critic — it is therefore not recommended to optimize one model and assume that the same values will hold for the other. Thus DDPG entails a significantly higher optimization burden than DQN.

Finally, we researched **Twin Delayed Deep Deterministic Policy Gradient** (TD3). TD3 is a newer algorithm than DDPG or DQN, and at the time of our research it had not yet been fully integrated with ReAgent. Therefore, we had to build our own training pipeline in PyTorch. TD3 is an actor-critic algorithm which allows for continuous action spaces, while promising improved learning speed and performance over DDPG [Fujimoto et al., 2018].

In order to investigate the viability of each algorithm, we trained and tested them using an offline dataset based on our purchased inventory. Each policy was trained on the same portion of the dataset, and then we ran our RTB simulator, using the policy to choose the viewability threshold on a regular basis. Final performance was assessed in the manner described in Section 7. The algorithms compared were: a DQN with an action space of 10 actions; a DQN with 20 actions; DDPG; and TD3. The reward was plotted as a function of the number of testing intervals.

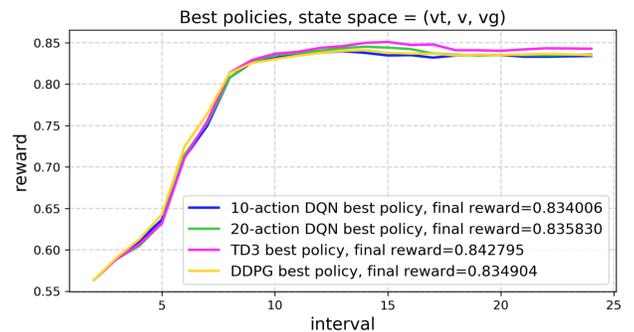

*Figure 4: Comparison of reward from various RL algorithms.*

We can observe that the TD3 policy had the highest performance of the four; the other three performed similarly with each other by the end of the test. Given that TD3 and DDPG both allow for a continuous action space, this obviates DQN as a primary solution. Thus, in general we would prefer TD3 to DDPG, given its superior performance and faster training time; however, the wider availability of DDPG may result in it being a compelling choice under some circumstances.

# 7. HYPERPARAMETER TUNING WITH BAYESIAN OPTIMIZATION

Training a deep reinforcement learning policy requires optimizing many hyperparameters simultaneously — including minibatch size, learning rate, number and size of hidden layers, and so on. With an actor/critic model we need to make these decisions for two separate networks, and in many cases the ideal configuration between the two networks is not the same.

With a small parameter space it would be possible to grid search all configurations, but that would not be realistic for this situation; first because of the number of possible configurations, and second because of the time to test each selection. The training and testing process of a single configuration — which will be described in further detail below — would typically require between several minutes to several hours. Therefore, any tractable approach must involve some stochastic exploration of the parameter space, followed by highly targeted exploitation.

Our solution was to use Bayesian optimization to optimize hyperparameters. This approach is well-suited for the problem, by starting out with a stochastic exploration process, followed by a more targeted exploration as observations are taken and the posterior distribution improves. We used the *BayesianOptimization* Python package [2], which automates most of the process, and can save its progress in case of interruptions. Policy training was performed using Facebook ReAgent with a DDPG algorithm. We used an AWS server with GPUs, which minimized training time and effectively eliminated the possibility of system downtime (which could be a significant issue when training on local machines).

Parameters optimized included:
- Learning rate for actor and critic
- Number of training epochs
- Minibatch size
- Gamma (Future reward discount rate)

Following a policy being trained, it was tested offline, using an RTB simulator as in Section 4.2. The simulator ran through a day of bidding, and used the trained policy to decide the viewability threshold on regular intervals. At the end of the simulated day of bidding, a single reward score *r* (Equation 1) was generated to assess how effectively the policy optimized viewability:

$$r = (1 - abs(v - vg))^2$$

Where *v* is measured viewability percentage, and *vg* is the viewability percentage goal. Because one possible edge case involves significantly increased viewability at the expense of not being able to spend the full budget, alternative reward metrics, including pacing and other KPIs, were tested as well. (In practice, these gave similar results to the simpler aforementioned reward, but would likely be useful for a more complex implementation with a larger state space.)

---
[2] https://github.com/fmfn/BayesianOptimization

The final reward value was passed back into the Bayesian optimization package, thus serving as an objective function by which the algorithm could evaluate each policy.

On the AWS server, each simulation run took roughly five minutes; the training was significantly influenced by factors such as minibatch size and learning rate, and generally varied between several minutes and an hour.

We found the Bayesian optimization approach to be very effective in navigating the large number of parameters involved in training an effective policy with a DDPG algorithm. In most cases it was not clear from the outset what values, or even ranges, would be ideal; Bayesian optimization was able to find effective parameters in a reasonable amount of time, and allowed us to avoid making selections on an arbitrary basis.

# 8. EXPERIMENTAL RESULTS

## 8.1 Offline results based on RTB simulator

We use a selection of historical campaign data to test the performance of multiple policies (this simulation process is discussed in section 4.2). None of the campaigns used in the testing of policies were used in the training;

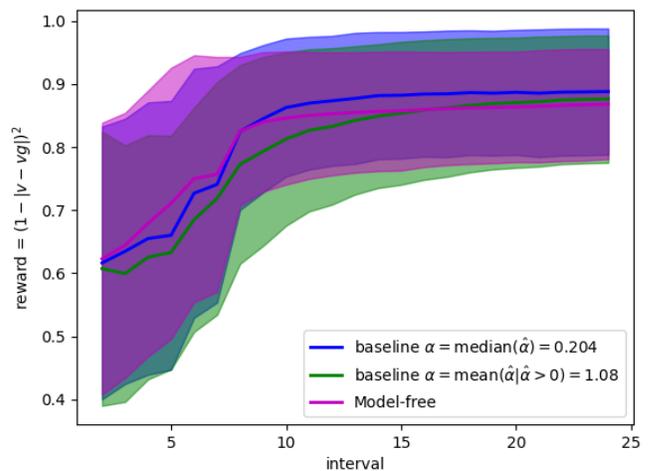

*Figure 5: Comparing baseline policy performance*

random rollouts for model-free policy training were selected from a different set of historical campaign data.

We first investigated which model-based baseline policy performed best. We experimented with two different settings of the baseline model parameter $\alpha$, $\alpha = median(\hat{\alpha}) = 0.204$, and $\alpha = mean(\{\hat{\alpha} \mid \hat{\alpha} > 0\}) = 1.08$.

As shown in Figure 5, the $\alpha = median(\hat{\alpha}) = 0.204$ model yielded consistently higher reward than that of the $\alpha = mean(\{\hat{\alpha} \mid \hat{\alpha} > 0\}) = 1.08$, although the difference decreased approaching the final interval.

While the model-free policy outperformed both baselines early on, over time the latter policies caught up, and by the final interval both model-based approaches yielded slightly more reward than that of model-free.

## 8.2 Online results

Our online tests were conducted against live traffic for several weeks. The control used the rule-based feedback loop method in Tashman et al. [2020], with a PID-based controller and an actuator applying adjustments on a 6-hour interval.

We observe in Figure 6 that the control required a significantly greater amount of time to reach its goal; this is a consequence of its rule-based feedback-control method only being able to approach the correct viewability threshold through slow adjustments rather than exact modeling. It is therefore designed to take actions conservatively, as extreme initial guesses could unacceptably impact delivery or campaign performance.

In contrast, the RL approach was able to aggressively raise the viewability threshold and bring viewability quickly to goal. Its initial actions were effective and delivery was not harmed. After the goal was raised on Day 26, the test was consistently closer to goal than the control.

However, after the goal was reduced, the test algorithm did not effectively respond and reduce the viewability threshold. In contrast, before the experiment was ended, the control group did begin to reduce viewability.

This outcome reflects the tradeoffs and potential problem for reinforcement learning in RTB. In cases in which the policy is effectively trained, it is capable of acting more quickly and precisely than that of a PID approach, and is able to adjust its response in nuanced ways according to the specifics of its environment. However, it is not highly robust to holes in training data, and its response to edge cases can be difficult to predict. In contrast, the PID approach behaved consistently and predictably throughout the test.

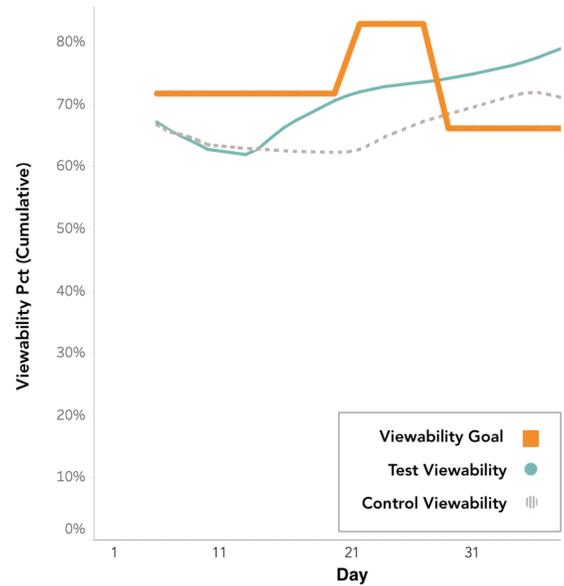

Figure 6: Comparison of RL control versus a *rule-based* PID-based feedback loop

## 9. CONCLUSIONS

Deep reinforcement learning is an extremely promising technique which has been applied to a variety of fields. Many problems in digital advertising are amenable to RL, given the commonality of large datasets and quick feedback loops. This paper surveyed various RL algorithms and their tradeoffs, presented methods for expanding training data into edge cases, discussed Bayesian hyperparameter optimization, and covered an assortment of live-traffic tests.

There are several areas in which future research would be beneficial. Training for a larger action space, including bid price multipliers, would be immensely useful for RTB. Training on a large array of environment variables could allow for better-informed decisions. Finally, an exploration of the newest RL algorithms, beyond the ones included here, would be highly valuable.

## ACKNOWLEDGEMENTS

The authors would like to thank Adam Cushner, Victor Seet, and Tobias Sutters for their contributions.